# A Comparative Study of Confidence Calibration in Deep Learning: From Computer Vision to Medical Imaging[*]

Riqiang Gao[a], Thomas Li[b], Yucheng Tang[c], Zhoubing Xu[e], Michael Kammer[d], Sanja L. Antic[d], Kim Sandler[d], Fabien Maldonado[d], Thomas A. Lasko[d] and Bennett Landman[a,b,c]

[a]Computer Science, Vanderbilt University, Nashville, TN, 37235, USA

[b]Biomedical Engineering, Vanderbilt University, Nashville, TN, 37235, USA

[c]Electrical and Computer Engineering, Vanderbilt University, Nashville, TN, 37235, USA

[d]Vanderbilt University Medical Center, Nashville, TN, 37235, USA

[e]Siemens Healthineers, Princeton, NJ, 08540, USA

## ARTICLE INFO

## ABSTRACT

Although deep learning prediction models have been successful in the discrimination of different classes, they can often suffer from poor calibration across challenging domains including healthcare. Moreover, the long-tail distribution poses great challenges in deep learning classification problems including clinical disease prediction. There are approaches proposed recently to calibrate deep prediction in computer vision, but there are no studies found to demonstrate how the representative models work in different challenging contexts. In this paper, we bridge the confidence calibration from computer vision to medical imaging with a comparative study of four high-impact calibration models. Our studies are conducted in different contexts (natural image classification and lung cancer risk estimation) including in balanced vs. imbalanced training sets and in computer vision vs. medical imaging. Our results support key findings: (1) We achieve new conclusions which are not studied under different learning contexts, e.g., combining two calibration models that both mitigate the overconfident prediction can lead to under-confident prediction, and simpler calibration models from computer vision domain tend to more generalizable to medical imaging. (2) We highlight the gap between general computer vision tasks and medical imaging prediction, e.g., calibration methods ideal for general computer vision tasks may in fact damage the calibration of medical imaging prediction. (3) We also reinforce previous conclusions in natural image classification settings. We believe that this study has merits to guide readers to choose calibration models and understand gaps between general computer vision and medical imaging domains. [1]

## 1. Introduction

Confidence calibration measures the agreement between the predicted probability and the true class prevalence (expected accuracy) [13]. Confidence calibration is important across domains, especially health-care, for trustworthy prediction [6, 21, 3]. A predictive model should know when its prediction is inaccurate. For example, in computer-aided diagnosis system, the model should inform its users, doctors, of the uncertainty around its prediction.

[1]Our implementation is publicly available at https://github.com/MASILab/ComparativeCalibration.

[*]This research was supported by NSF CAREER 1452485, R01 EB017230 and R01 CA253923. This study was supported in part by U01 CA196405 to Massion.

riqiang.gao@vanderbilt.edu (R. Gao)

riqianggao.github.io (R. Gao); https://my.vanderbilt.edu/masi/ (R. Gao)

ORCID(s): 0000-0002-8729-1941 (R. Gao)

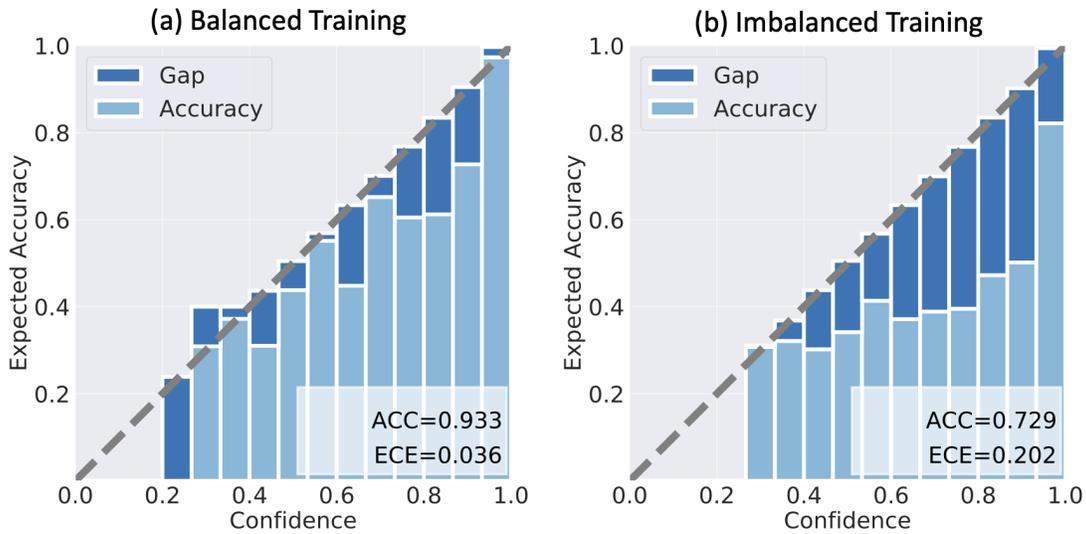

**Figure 1:** The reliability diagrams on the same test set from balanced and imbalanced training (more details can be found in experiments section) on CIFAR10. Both the performances of class discrimination reflected by accuracy (ACC, larger is better), and confidence calibration reflected by expected calibration error (ECE, smaller is better) are worse under the imbalanced training.

Due to the availability of large-scale datasets and powerful computing resources, recent development of deep learning models has dramatically improved the discrimination of prediction models [26, 16, 36]. However, improvements in model calibration are less impressive. Deep learning models with poor calibration primarily result from overconfidence in the training data [43, 23]. Additionally, hyperparameters like model depth, weight decay and batch normalization influence final model calibration [13].

Apart from general model development challenges, prediction with medical imaging has its own obstacles to achieving desirable calibration. First, the number of training samples in medical imaging are usually much smaller than that in computer vision. For example, the ImageNet [9] has over 1 million samples with high-quality labels, while NLST [35], as one of the largest lung cancer datasets, has only around 5 thousand patients with high-quality label (i.e., biopsy confirmed cancer diagnosis within 1 year). Second, the available training data for clinical diagnosis are usually highly unbalanced. For example, the national lung screening trial (NLST) includes over 23,000 patients in the chest CT arm, but only about 1000 patients (4%) were ultimately diagnosed with lung cancer [35]. Even though great progress has been made in deep learning with publicly available well-balanced large-scale datasets such as ImageNet [5], CelebA [31], COCO [30], model discrimination and calibration can be challenging in the imbalanced training [41, 22, 40, 47].

As shown in Figure 1, imbalanced training sets damage the class discrimination and confidence calibration. Though studies have been conducted to improve the discrimination in re-sampling [4, 5, 19] or re-weighting contexts [44, 17, 8, 29], the calibration analyses on imbalanced training is relatively understudied, especially in clinical contexts. In addition, the medical imaging may be weakly labeled. For example, in many lung cancer imaging datasets, the label is on the level of the CT scan rather than on the pulmonary nodule of interest. Additional approaches are needed to handle these weak labels (see Section 5 for our example using multi-instance learning). These challenges make the training of the deep learning models not ideal mode and can affect the confidence calibration. In summary, there are numerous challenges in calibration that are unique to the medical imaging domain.

Confidence calibration is commonly performed in either one or two stages. The one-stage approach attempts to simultaneously optimize the discrimination and calibration of a model as part of training. The two-stage approaches separate representation learning and classifier training for better performance [47].

The one-stage approach of mixup [45] includes a regularization technique for deep learning that linearly interpolates the input and label. Mukhoti et al. propose the use of focal loss [29] can train better calibrated deep learning models from theoretical and empirical perspectives [32]. Researchers [42, 20] showed that training with mixup can significantly improve the model calibration. Some methods improving model calibration is under the uncertainty estimation context. Based on the assumption that a well-calibrated model should be accurate when it is certain (with high confidence) and be uncertain (with low confidence) when it is inaccurate, Krishnan et al. [24] introduced the differentiable accuracy versus uncertainty calibration (AvUC) loss function which was shown to optimize models that have well-calibrated uncertainties. Karandikar et al. [23] introduce differentiable soft calibration objectives (e.g., soft version of AvUC) to improve calibration.

There are also calibration models in a two-stage framework. One of the earliest works on calibrating deep learning models is discussed in [13], which offers calibration insights into neural network learning and provide a simple but effective way (i.e., temperature scaling) to calibrate models. Label smoothing [39] is another widely applied regularization technique that helps avoid overfitting during training. Muller et al. [33] introduced the use of label smoothing as a method to increase the confidence calibration. Zhong et al. [47] incorporate a label-aware smoothing in the calibration model for long-tail learning in the second stage. The techniques of Krishnan et al. [24] and Karandikar et al. [23] have been applied in the second stage as well.

Even though promising calibration results have been demonstrated in previous arts, most of them are under standardized (e.g., large-scale training, balanced training) or simulated classification tasks. The imbalanced-training challenges especially in clinical contexts have not been well explored. While approaches for calibrating deep learning models have been proposed recently, there has not been a comprehensive study of calibration models applied in different challenging contexts, especially clinical disease diagnosis. As a representative example, lung cancer has the highest mortality across all cancers [38, 37]. Deep learning models have been widely applied to diagnose lung cancer using computed tomography (CT) images with satisfactory discrimination [2, 28, 11, 12], however, confidence calibration of the prediction models is lacked.

In this work, we conduct a comparative study of four leading one- and two-stage calibration models in different contexts and domains for a total of 12 experiments for comparison. Specifically we compare calibration models when used in a balanced multi-class classification task with large-scale training set (i.e., original CIFAR10 [25]), a simulated imbalanced training classification context with large-scale datasets (i.e., long tail CIFAR10 following [47]). and a clinical prediction task (i.e., lung cancer diagnosis) with CT images including internal- and external- validation with three data cohorts. Our study seeks to help readers to understand big picture of calibration models with four leading approaches and demonstrate their varying effectiveness across different domains.

The rest of manuscript are organized as follows: Section 2 introduces and discusses the selected four calibration methods across Stage 1 and 2, including their high impact notes. Section 4 shows our experiments in computer vision including balanced and imbalanced training. Our clinical prediction task (i.e., lung cancer diagnosis) are demonstrated in Section 5. Finally, we conclude and discuss our work with take-home messages in Section 6.

## 2. Calibration Methods

In this section, we introduce the used representative approaches for calibration in this paper. Our motivation to choose the approaches is (1) well-known and widely used, (2) cover multiple categories such as one- vs. two- stages. With the 4 selected approaches and the baseline cross-entropy loss, we have 12 combinations in total for comparison.

**Baseline: cross-entropy loss (CEL)**. CEL is also known as negative log likelihood (NLL), which measures the quality of a probabilistic model [15]. CEL is commonly used in deep learning for classification [27], and is defined as followed:

$$CEL(y, p) = \sum_{k=1}^{K} -y_k \log(p_k),$$

where $p_k$ is the predicted probability of class $k$. $y_k = 1$ for the correct class and $y_k = 0$ for the rest classes.

**Focal Loss** [29]. Focal loss was first proposed for dense object detection to deal with class imbalance problems. Let's denote the predicted probability corresponding to target class $y_t$ as $p_t$, which ranges from 0 to 1. To prevent easily classified negatives dominating the loss and gradient, focal loss includes a modulating factor $(1 - p_t)^\gamma$ to cross entropy loss, where $\gamma$ is a positive number as the exponent. In practice, the focal loss can be adopted in the $\alpha$-balanced form:

$$FL(p_t) = -\alpha_t (1 - p_t)^\gamma \log(p_t)$$

where $\alpha_t$ is the hyper-parameter associated to class $y_t$.

When $p_t$ is close to 1, which indicates the sample can be easily classified, the modulating factor $(1 - p_t)^\gamma$ would be small. Thus, compare with CEL, Focal loss has less penalty on easy samples. This reduces the overconfidence of training.

*Note: Focal loss was proposed in ICCV 2017 (top-tier, H5 Index: 172) [29] and won the best paper, then was studied for calibration in NeurIPS 2020 (top-tier, H5 Index: 192) [32]. Until March 8, 2022, It has gotten 12,063 citations.*

**Label-Aware Smoothing** [33]. Label smoothing (LS) [39, 33] applies a transformation to ground truth labels such that $y_k^{LS} = y_k(1 - \alpha) + \alpha/K$. where $\alpha$ is the smoothing factor to relief the overconfidence in training. To improve calibration for long-tailed recognition, Zhong et al. [47] proposed the Label-aware Smoothing (LAS) where the smoothing factor of class $y$ depends on class number $N_y$, as below:

$$y_k^{LAS} = y_k(1 - f(N_y)) + \frac{f(N_y)}{k},$$

where large $N_y$ has a larger smoothing factor $f(N_y)$.

Thus, the ground truth label in LAS is no longer a one-hot vector. Consider a 3-class classification task with training sizes of 10000, 5000, 1000 for class 1, 2, and 3 respectively. The one-hot representation of ground truth for CEL is class 0: [1, 0, 0], class 1: [0, 1, 0], class 2: [0, 0, 1]. The LS representation can be class 0: [0.8, 0.1, 0.1], class 1: [0.1, 0.8, 0.1], class 2: [0.1, 0.1, 0.8], and the LAS representation can be class 0: [0.8, 0.1, 0.1], class 1: [0.05, 0.9, 0.05], class 2: [0.02, 0.02, 0.96]. Under this situation, the head classes (i.e., the classes with more samples) has larger regularization to avoid overfitting. Following [47], the LAS is applied in the second stage.

*Note: Label-smoothing was proposed and studied in CVPR 2016 (top-tier, H5 Index: 301) [39] and specially studied in NeurIPS 2019 (top-tier, H5 Index: 192) and ICML 2021 (top-tier, H5 Index 151) [33], then was studied for calibration in CVPR 2021 (top-tier, H5 Index: 301) [47]. Until March 8, 2022, it has gotten 19,149 citations.*

**Mixup** [45]. Mixup strategy is motivated by the Vicinal Risk Minimization [7]. As in [45], the vicinal points $(\tilde{x}, \tilde{y})$ are generated according to the following rules:

$$x = \lambda x_i + (1 - \lambda)x_j \\ y = \lambda y_i + (1 - \lambda)y_j \tag{4}$$

where $x_i$ and $x_j$ are randomly selected sample pairs. $y_i$ and $y_j$ are related one-hot labels of $x_i$ and $x_j$. $\lambda \in [0, 1]$ and $\lambda \sim beta(\alpha, \alpha)$ for $\alpha \sim (0, \infty)$. The mixup strategy assumes that linear interpolation of inputs should linear interpolation of outputs. Mixup training has been shown to contribute to the robustness and generation of models [46].

*Note: mixup was proposed in ICLR 2017 (top-tier, H5 Index: 166) [45], then was studied for calibration in NeurIPS 2019 (top-tier, H5 Index: 192) [42], and theoretically explained in ICLR 2021 (top-tier, H5 Index: 166) [46]. Until March 8, 2022, it has gotten 3426 citations.*

**Temperature Scaling** [13]. Platt scaling is a parametric approach for model calibration, which learns two parameters a,b to achieve a new prediction $q_i = \sigma(a \cdot z_i + b)$ given the original prediction $z_i$. Temperature scaling is a simple and straightforward extension of Platt scaling, which only has one learnable parameter. In the multi-class classification context, given the logit vector $z_i$, the new prediction is obtained as follows:

$$\hat{q}_i = \max_k \sigma_{SM}(\frac{z_i}{T})^{(k)} \tag{5}$$

where the parameter $T$ is called temperature and $\sigma_{SM}$ is the Softmax function. Since both plat scaling and temperature scaling do not change the ranking of predictions, so they do not affect the discrimination of the overall model.

*Note: temperature scaling was studied for calibration in ICML 2017 (top-tier, H5 Index: 192) [13]. Until March 8, 2022, it has gotten 2237 citations.*

## 3. Evaluation Methods

In this section, we introduce the quantitative metrics and illustration diagrams to evaluate the model's calibration as well as discrimination.

**Expected Calibration Error (ECE)**. ECE [34] is one of the most popular metrics to measure calibration performance. There are two main steps to compute ECE: (1) dividing the prediction value space into equal-space bins, (2) calculating the weighted average of the difference of accuracy and confidence. The ECE is defined as:

$$ECE = \sum_{m=1}^{M} \frac{|B_m|}{n} |acc(B_m) - conf(B_m)| \tag{6}$$

where the $n$ is the number of samples, $B_m$ is the m-th bin.

**Accuracy (ACC)**. Accuracy is a metric that measures the fraction of model predictions get right.

$$ACC = \sum_{i=1}^{N} \mathbf{1}\hat{y}_i == y_i)/N \tag{7}$$

where $\hat{y}_i$ and $y_i$ is the predicted label and ground truth for sample $i$, respectively. $N$ is the number of samples. ACC is used for evaluating the discrimination of multi-class classification.

**Area Under the ROC Curve (AUC)**. ROC curve (receiver operating characteristic curve) is a figure showing the performance of a prediction/classification model at all thresholds [10]. The curve plots true positive rate as y-axis and false positive rate as the x-axis. The AUC represents the area of the two-dimensional region under the whole ROC curve from (0,0) to (1,1). AUC is used for evaluating the discrimination of binary classification.

**Reliability Diagrams**. Reliability diagrams [14] are the figures of the predicted probabilities (confidence) and the observed frequency (expected frequency), which shows the frequency of predicted probabilities happened in practice (i.e., confidence calibration). Examples can be found in Figure 1, 2 and 4.

**Histograms of Predicted Probability**. The histogram shows the distribution of predicted probability of the corresponding class. The histograms indicate how confident (high or low) of the model in the population. Integrated with reliability diagrams, the overconfidence or under-confidence can be inferred. These are depicted in Figure 3 and 5.

## 4. Natural Image Classification: CIFAR10

### 4.1. Data Introduction

The CIFAR10 consists of 60,000 tiny images evenly from 10 classes. There are originally 50,000 samples in the training set and 10,000 samples in the test set. We further split the original training set into training/validation splits with the ratio of 9:1. The validation set is used for tuning hyper-parameters and selecting epoch number.

The original CIFAR10 [25] dataset is balanced, and we term it as CIFAR10-Ori in this chapter. Following the setting of [47], we create a variant of CIFAR10-Ori with long-tail distribution called CIFAR10-LT with the imbalance factor (IF) 0.01 for the training set. The number of samples in class $i$ is defined as $N_i = N_{ori} \cdot IF^{i/9}$, where $N_{ori}$ is the sample number in CIFAR10-Ori. The validation and test sets of CIFAR10-LT keep the same as CIFAR10-Ori.

### 4.2. Implementation

We utilize the ResNet-32 [16] as our backbone network, as was previously studied in [47]. The training process has up to 2 stages. For Stage 1, the feature extraction model ResNet-32 is trained from scratch with Focal loss or cross-entropy loss (CEL), and we also have the comparison of with mixup and without mixup. For Stage 2, the feature extraction model is fixed. In this case, only the classifier (last layer) was trained, or the logits were scaled. We include the Label-aware Smoothing (LAS) [47] and Temperature Scaling (TS) [13] for comparison. The experiments are conducted on both CIFAR10-Ori and CIFAR10-LT.

We use the SGD optimizer with a momentum of 0.9 and a base learning rate (BLR) of 0.1. In the first five epochs, the learning rate is $BLR * epoch_{index}/5$. The learning rate is divided by 10 at 150th and 250th epoch. The max number of epochs is 300. The batch size is set to 128 and the weight decay is 5e-4. Our experiments are conducted with PyTorch 1.6. The hyper-parameter $\alpha$ in mixup is set to 1 for CIFAR10-LT and 0.2 for CIFAR10-Ori. The hyper-parameters in focal loss are $\alpha = 1, \gamma = 2$. The number of bins when computing the ECE is 15.

### 4.3. Results and Their Analyses

Herein, we first present the easier setting of balanced training (i.e., CIFAR10-Ori) and then the harder setting of imbalanced training (i.e., CIFAR10-LT).

**Balanced CIFAR10-Ori Results**.

Table 1, Figure 2, and Figure 3 show the experimental results on CIFAR10-Ori. From the results of Stage 1, the discrimination performance (ACC) was similar across four settings. Compared with the baseline CEL, the strategy of mixup slightly improved the ACC while improve calibration performance with a clear margin (ECE: 3.60%->2.86%). In contrast, replacing the CEL with Focal loss slightly decreased the ACC and improved the ECE with a large margin (ECE: 3.60%->1.22%). When adding both the mixup and Focal loss, the calibration performance became distinctly worse.

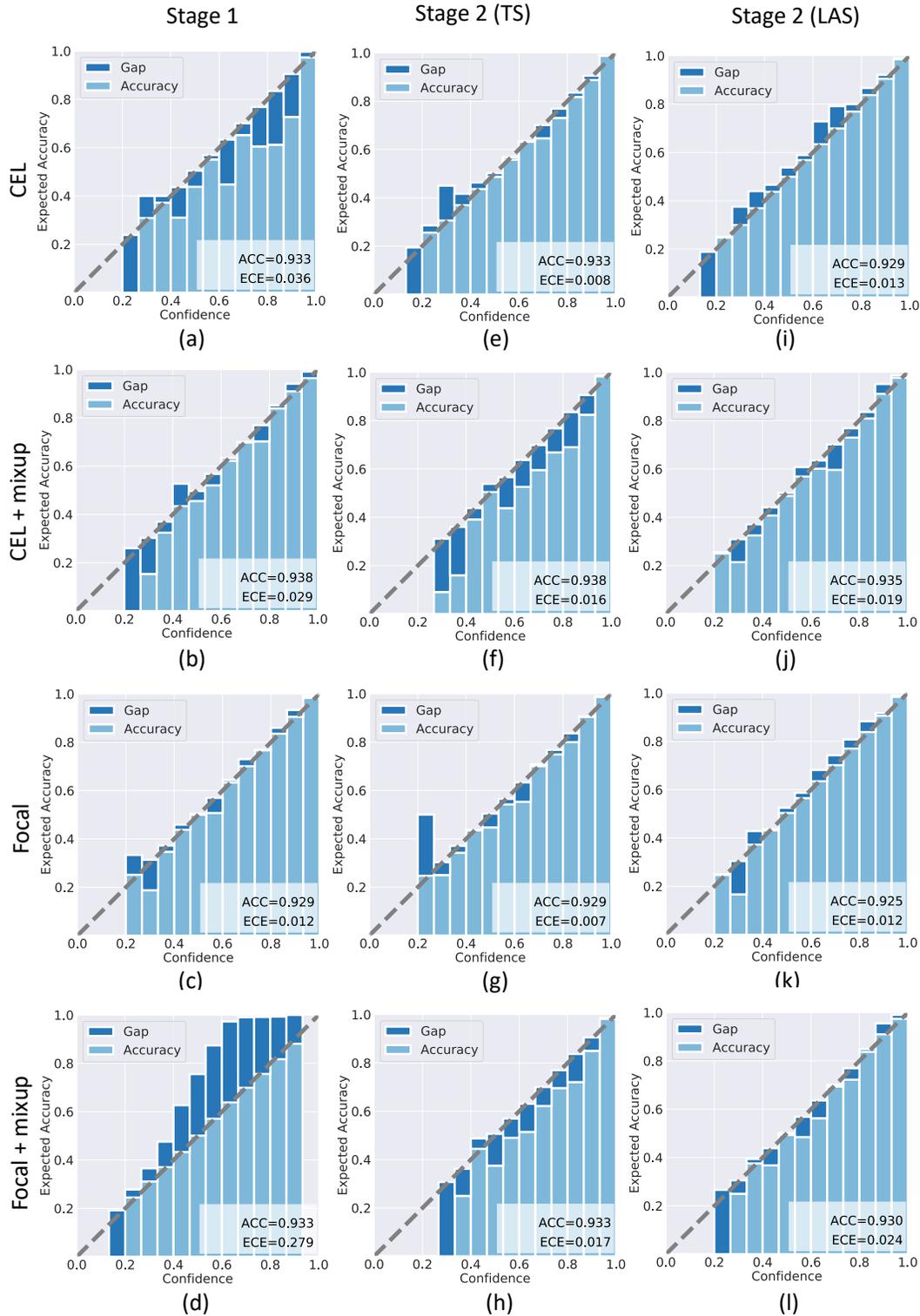

**Figure 2:** Reliability Diagrams of experiments on CIFAR10-Ori. This figure corresponds to the results in Table 1. Note that if the Accuracy is larger than the Expected Accuracy, the part of light blue bar would be covered by dark blue bar.

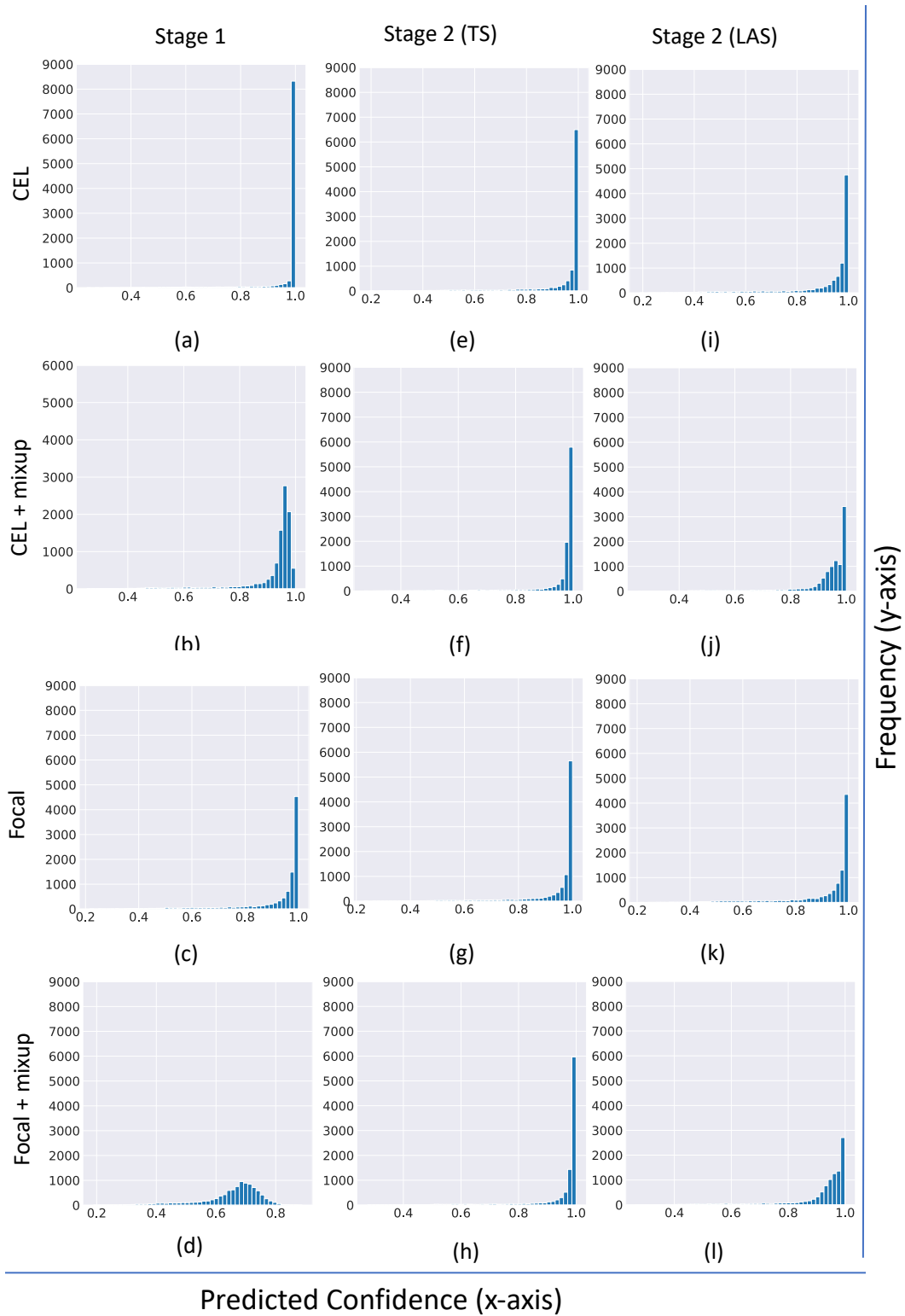

**Figure 3:** Histograms of experiments on CIFAR10-Ori (x-axis: predicted confidence, y-axis: sample count). This figure corresponds to the results in Table 1. The histogram shows how confident the prediction is, and when integrated with reliability diagrams, we can inference if the prediction is overconfident or underconfident.

**Figure 4:** Reliability Diagrams of experiments on CIFAR10-LT. This figure corresponds to the results in Table 2. Note that if the Accuracy is larger than the Expected Accuracy, the part of light blue bar would be covered by dark blue bar.

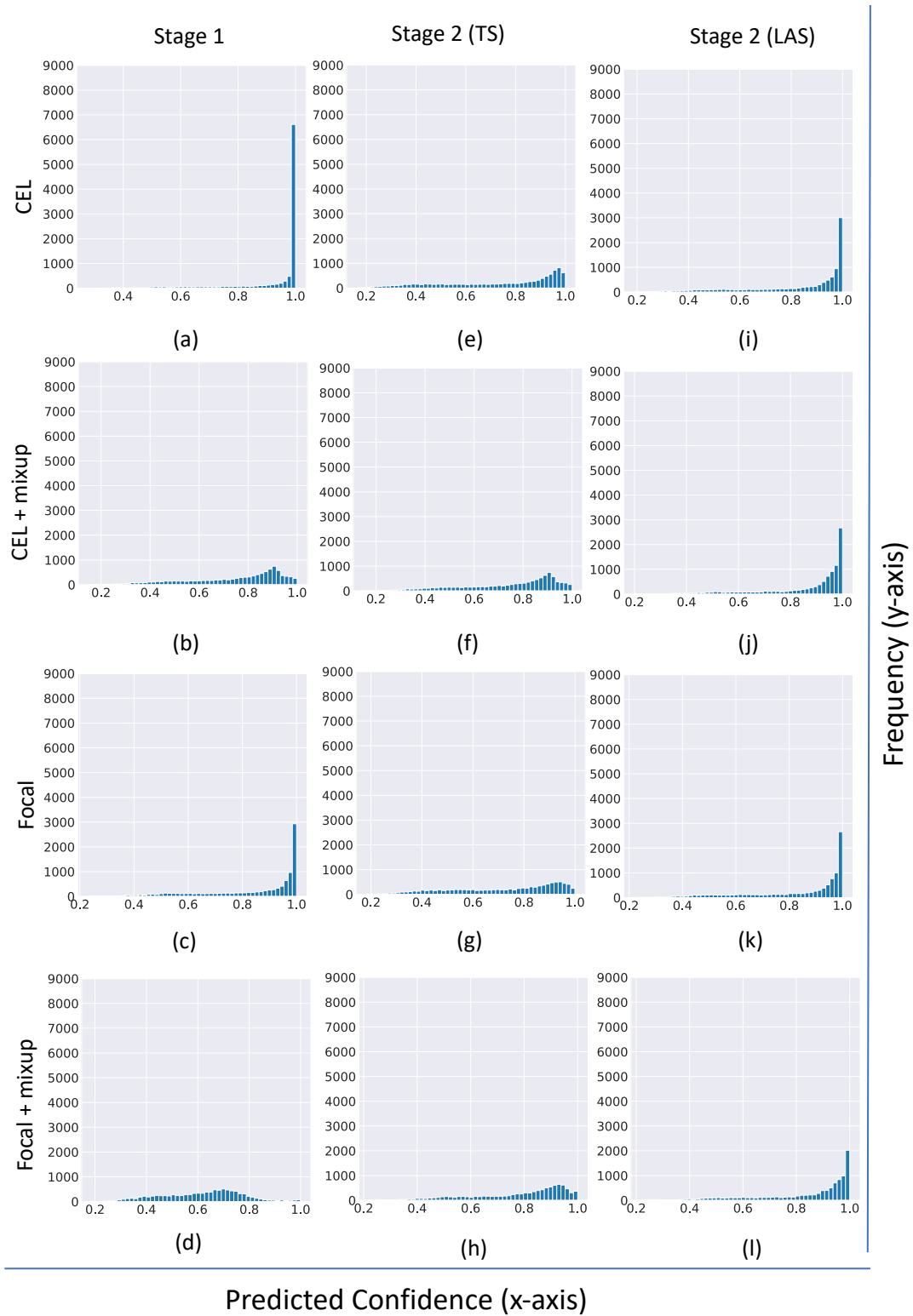

**Figure 5:** Histograms of experiments on CIFAR10-LT (x-axis: predicted confidence, y-axis: sample count). This figure corresponds to the results in Table 2. The histogram shows how confident the prediction is, and when integrated with reliability diagrams, we can inference if the prediction is overconfident or underconfident.

Comparative Study of Calibration Analysis

| Stage 1 Model | Stage 1 | | Stage 2 (temp) | | Stage 2 (LAS) | |
|---|---|---|---|---|---|---|
| | ACC (↑) | ECE (↓) | ACC(↑) | ECE (↓) | ACC(↑) | ECE (↓) |
| CEL | 93.32 | 3.60 | 93.32 | 0.75 | 92.86 | 1.32 |
| CEL + Mixup | 93.83 | 2.86 | 93.83 | 1.57 | 93.54 | 1.87 |
| Focal | 92.95 | 1.22 | 92.95 | 0.65 | 92.55 | 1.29 |
| Focal + Mixup | 93.26 | 27.92 | 93.26 | 1.66 | 93.00 | 2.40 |

**Table 1**
Test results of training with CIFAR10-Ori. (↑) represents larger number is better. (↓) represents smaller number is better.

When applying the temperature scaling (TS) in Stage 2 for recalibration, the class discrimination performance did not change, while the confidence calibration saw a large improvement across all the four Stage-1 settings. When using the label aware smoothing (LAS) [47], the ACCs were slightly changed. Generally, LAS improved calibration performance in Stage 2, but less so than TS.

**Balanced CIFAR10-Ori Analyses**.

In Stage-1 comparison, as indicated in Figure 2 and 3, when trained by CEL, the predicted risks are highly concentrated near 1 (i.e., highly confident). Here, mixup and focal loss both individually reduced the overconfidence of CEL, which improves calibration. However, when mixup and focal loss were applied together, the prediction became underconfident, which hurt calibration.

As for class discrimination, there is no clear impact found (ACC changes < 1%) from mixup and focal loss. As expected, temperature scaling did not change discrimination across all experiments as it only scales the logits and will not change the ranking of predicted risks from different classes. As the second stage post-processing methods, both LAS and temperature scaling work well in confidence re-calibration.

In conclusion, all the selected approaches improved calibration except the combination of focal loss + mixup in the setting of training a classifier on a large-scale balanced dataset.

**Imbalanced CIFAR10-LT Results**.

The experiments of training with CIFAR10-LT are shown in Table 2. Compared with CIFAR10-Ori, the discrimination performance was lower by approximately ∼ 20% ACC.

In Stage 1, mixup greatly improved the calibration performance (ECE) and slightly improved the discrimination performance (ACC). The Focal loss improved the CEL of the baseline model but decreased the ECE when used with mixup (i.e., ECE: CEL + mixup (1.72) vs. Focal Loss + mixup (11.48)). The LAS improved calibration and discrimination except in the setting of CEL + mixup.

**Imbalanced CIFAR10-LT Analyses**.

The limited diversity of classes in the imbalanced training setting is widely known to limit a model's generalizability, decreased discrimination, and possibly lead to overfitting of overrepresented classes.

Like the analyses of CIFAR10-Ori, mixup, Focal loss and the second stage model LAS compensate for model overconfidence, which improved calibration performance. While all the selected approaches positively impacted performance in this simulated long-tail classification, LAS resulted in the best discrimination and the TS resulted in the best calibration.

In contrast to the CIFAR10-Ori setting, "Focal + mixup" had better calibration performance than the baseline "CEL". However, individually they out performed the combination of the two due to under-confidence predictions when combining Focal loss and mixup.

| Stage 1 Model | Stage 1 | | Stage 2 (temp) | | Stage 2 (LAS) | |
|---|---|---|---|---|---|---|
| | ACC (↑) | ECE (↓) | ACC(↑) | ECE (↓) | ACC(↑) | ECE (↓) |
| CEL | 72.90 | 20.20 | 72.90 | 2.02 | 77.19 | 6.64 |
| CEL + Mixup | 74.10 | 1.72 | 74.10 | 1.72 | 81.54 | 5.18 |
| Focal | 72.18 | 12.22 | 72.18 | 2.32 | 76.83 | 7.19 |
| Focal + Mixup | 72.93 | 11.48 | 72.93 | 5.37 | 80.82 | 3.69 |

**Table 2**

Test results of training with CIFAR10-LT. (↑) represents larger number is better. (↓) represents smaller number is better.

## 5. Medical Imaging: Lung Cancer Diagnosis

### 5.1. Data Introduction

We have included three datasets in this experiment: the national lung screening trial (NLST) [35], and two cohorts, UPMC and UCD, from MCL [1].

**NLST**. The NLST is a randomized controlled nested case:control clinical trial for lung cancer detection that was designed to evaluate whether the screening with CT reduces mortality compared to screening with X-ray. About 54,000 participants across 33 centers were enrolled between 2002 and 2004. It represents one of the largest lung CT datasets available for research. This study used subjects that 1) have a tissue-based diagnosis, and 2) have a diagnosis within 1 years of the last scan for cancer case. There are in total 606 cancer subjects out of 5344 subjects. Similar to [12], the selected subjects are all high-risk patients (all received biopsies) and the distinction between cancer / non-cancer in our cohort is more challenging than a cancer classification task on the entire NLST cohort. Our dataset was randomly split into five even folds. Four folds were used for training and one-fold for validation.

**UPMC and UCD**. Unlike the NLST which is a nested case:control screening cohort with low cancer rate, the UPMC and UCD datasets are matched case:control studies, with a mix of screening and incidentally detected lung cancers, and have much higher cancer rate (∼ 0.5). There are 78 cancer cases out of 155 patients in UPMC and 52 cancer cases out of 96 patients in UCD.

### 5.2. Implementation

We have included one calibration method (i.e., focal loss) in Stage 1 and two calibration methods (i.e., temperature scaling and LAS) in Stage 2 for a total of six combinations for comparison. Because the mixup strategy is designed for single image classification and the lung cancer diagnosis experiment is a multi-instance learning problem, we did not evaluate mixup in the lung cancer diagnosis experiment.

The neural network backbone is motivated by the image path of our previous work [12], as shown in Figure 6.

We first apply the preprocessing steps and nodule detection model of [44] to raw CT image data. Suggested by [28], the top five confidence nodule proposals are enough to cover all the nodules. To speed the training/test process, we use 2D images rather than 3D. Axial/coronal/sagittal directions of the nodule proposal are formulated as three channel data with the dimension of $(3 \times 128 \times 128)$. We use a ResNet-18 backbone to extract image features of each nodule proposal. Then, the image features of the five nodule proposals are transferred to a single image feature vector with an attention-based multi-instance learning layer [18].

We use the SGD optimizer with the momentum of 0.9 and a learning rate of 0.005. The max number of epochs is 100 and final model is used for testing as we empirically notice that the training is converged after 100th epoch in our setting. The batch size is set to 128 and the weight decay is 1e-4. Our experiments are conducted with PyTorch 1.6. The hyper-parameters in focal loss are: $\alpha = 1, \gamma = 2$. The number of bins when computing the ECE is 10.

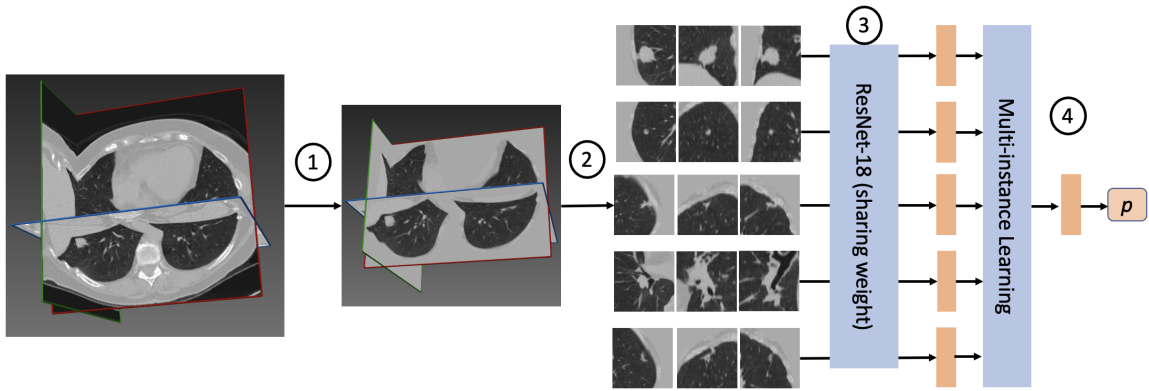

**Figure 6:** The framework for lung cancer diagnosis with CT image. The steps 1-4 are labeled in the figure: (1) image preprocessing: extracted lung regions and remove the unwanted tissue background. The raw CT and preprocessed image are 3D data. (2) nodule detection: locate nodule proposals with a pre-trained deep learning model. Validated by Liao model, proposals with top five regions are sufficient to detect cover possible nodules. (3) Feature extraction: extract high-level representations given the top five proposals in ResNet-18. (4) Feature combination and prediction: integrate features from nodule proposals with a multi-instance learning module as a single image feature and then get a cancer prediction.

| Stage 1 Model | Stage 1 | | Stage 2 (TS) | | Stage 2 (LAS) | |
|---|---|---|---|---|---|---|
| | ACC (↑) | ECE (↓) | ACC(↑) | ECE (↓) | ACC(↑) | ECE (↓) |
| CEL | 90.71 | 1.72 | 90.71 | 1.72 | 90.79 | 7.43 |
| Focal | 91.53 | 16.16 | 91.53 | 1.54 | 91.54 | 7.39 |

**Table 3**
Internal-validation results on NLST (%). TS represents temperature scaling and LAS represents label-aware smoothing.

| Stage 1 Model | Stage 1 | | Stage 2 (TS) | | Stage 2 (LAS) | |
|---|---|---|---|---|---|---|
| | ACC (↑) | ECE (↓) | ACC(↑) | ECE (↓) | ACC(↑) | ECE (↓) |
| CEL | 74.50 | 10.10 | 74.50 | 6.73 | 74.78 | 31.03 |
| Focal | 75.20 | 11.13 | 75.20 | 13.50 | 75.05 | 28.37 |

**Table 4**
External validation results on UCD. TS represents temperature scaling and LAS represents label-aware smoothing.

| Stage 1 Model | Stage 1 | | Stage 2 (TS) | | Stage 2 (LAS) | |
|---|---|---|---|---|---|---|
| | ACC (↑) | ECE (↓) | ACC(↑) | ECE (↓) | ACC(↑) | ECE (↓) |
| CEL | 83.50 | 12.60 | 83.50 | 6.53 | 83.70 | 24.23 |
| Focal | 87.20 | 10.50 | 87.20 | 7.97 | 87.20 | 19.89 |

**Table 5**
External validation results on UPMC. TS represents temperature scaling and LAS represents label-aware smoothing.

### 5.3. Experimental Results

The results on NLST, UCD, and UPMC are shown Table 3, Table 4, and Table 5, respectively. The Focal loss has improved discrimination over CEL, which has observed across all three datasets (e.g., 91.53% vs. 90.71% in NLST, 87.20% vs. 83.50% in UPMC). The temperature scaling generally improves the calibration performance across the losses of CEL and Focal. In most situations, the LAS does not improve the calibration performance but can improve the discrimination minorly.

## 5.4. Result Analyses

In general, Temperature scaling, as a simple post-processing, improved the calibration performance across our experiments. Interestingly, approaches that worked well for computer vision tasks failed to improve performance by the same margin in medical prediction. For example, LAS greatly improved both class discrimination and confidence calibration in CIFAR10-LT, while it resulted in a decrease in discrimination and calibration of lung cancer diagnosis. We believe the differences between natural image classification and medical imaging detailed below may explain the failure of some calibration models in lung cancer diagnosis.

*First*, the lung cancer diagnosis is based on a multi-instance learning (MIL) pipeline. The training label is assigned on the whole CT image (i.e., the "sample" of MIL) level rather than individual pulmonary nodules (i.e., the "instance" of MIL) level. Developing models on less granular, image-level labels is known to be more challenging than doing so on fine-grain, object-level labels.

*Second*, most calibration models in computer vision are developed and evaluated under well-defined and standardized contexts. For example, "cat" and "bird" images from CIFAR10 can be easily distinguished if high-level features are extracted effectively. While in lung cancer diagnosis, malignant and benign nodules can be hard to distinguished even by radiologists. These difficulties yield poor discrimination and calibration in clinical contexts.

*Third*, the number of training samples in CIFAR10 experiments is larger than that of lung cancer diagnosis. As deep learning models are data hungry, small training data size can negatively impact discrimination and calibration.

## 6. Discussion

Confidence calibration is important for prediction models, especially in the context of clinical diagnosis. It is an essential consideration for the ethical discussions surrounding implementation of deep learning in clinical decision making. Accurate confidence calibration is an essential to informing physicians when a prediction is uncertain so that they may question the prediction in the context of their clinical judgment. Even though deep learning has been successful in class discrimination and calibration studies have been conducted on standard or simulated datasets, the calibration on the challenging clinical decision-making contexts with deep learning models is understudied.

In this paper, we conducted a comparative study on confidence calibration and class discrimination with four high-impact calibration approaches. Those four approaches include one-stage and two-stage methods, and we have 12 method combinations for comparison. We evaluated these approaches in three settings: balanced large-scale training set on natural image classification (i.e., CIFAR10-Ori), large-scale imbalanced training (i.e., CIFAR10-LT), and lung cancer diagnosis with cross- and external- validations using three cohorts.

Our work is the first to evaluate popular learning methods for domain-based calibration differences between natural and medical images. We have found the following:

- Balanced training with CEL leads to overconfident predictions large calibration error (as shown in Figure 3(a), the prediction prevalence near 1 is high). Both the mixup and focal loss improves the calibration performance by reducing the overconfidence of the predictions (as shown in Figure 3(b), 3(c)). However, combining focal loss and mixup resulted in under confident predictions which damaged the calibration (see Figure 3(d)).

- Stage-2 model achieves similar calibration performance (Table 1 and 2) and risk distribution (Figure 3 and 5) across 4 stage-1 models (this is more obvious in temperature scaling).

- The calibration performance of both focal loss and label-aware smoothing was worse on medical images compared to natural images, suggesting different learning contexts (e.g., large-scale v.s. smaller-scale training sets, classical learning v.s. multi-instance learning, fundamental difference in image structure) can lead to different performance of calibration methods. We have conducted discussions (in Section 5.4) to analyze the gaps between computer vision and medical imaging domains.

- The simple approach of temperature scaling improved the calibration performance in both domains, suggesting that simpler approaches may be more robust to domain differences. Overall, in CIFAR10, LAS can improve discrimination and calibration in long-tail settings. However, LAS can decrease the model's discrimination and calibration.

- In agreement with previous work [13, 47, 32, 42], temperature scaling maintained the discrimination and improved calibration in all scenarios, including balanced- vs. imbalanced, nature image vs. medical image et cetera; focal loss, mixup, and label-aware smoothing improve the calibration of natural image classification in both balanced and imbalanced setting; and label-aware smoothing further greatly improved the discrimination in CIFAR10-LT.

In summary, we address the confidence calibration challenge of prediction models with empirical validations and are the first to highlight the gaps of calibration between computer vision and clinical diagnosis. We believe our study has insights to better understand calibration models and knowledge transfer between domains. Our code is publicly available at: `https://github.com/MASILab/ComparativeCalibration`.

## 7. acknowledgement

This research was supported by NSF CAREER 1452485, R01 EB017230 and R01 CA253923. This study was supported in part by U01 CA196405 to Massion. This project was supported in part by the National Center for Research Resources, Grant UL1 RR024975-01, and is now at the National Center for Advancing Translational Sciences, Grant 2 UL1 TR000445-06. This study was funded in part by the Martineau Innovation Fund Grant through the Vanderbilt-Ingram Cancer Center Thoracic Working Group and NCI Early Detection Research Network 2U01CA152662 to PPM.